\documentclass[10pt,twocolumn,letterpaper]{article}

\usepackage{cvpr}
\usepackage{times}
\usepackage{epsfig}
\usepackage{graphicx}
\usepackage{amsmath}
\usepackage{amssymb}
\usepackage{multirow}
\usepackage{textcomp}
\usepackage{wasysym}
\usepackage{subfigure}
\usepackage{xcolor}
\usepackage{float}
\usepackage{textcomp}
\usepackage{dsfont}
\usepackage{color}
\usepackage{colortbl}

\usepackage[pagebackref=true,breaklinks=true,letterpaper=true,colorlinks,bookmarks=false]{hyperref}

\cvprfinalcopy 

\def\cvprPaperID{475} 

\makeatletter
\@newctr{footnote}[page]
\makeatother

\ifcvprfinal\fi
\newfloat{figtab}{htb}{fgtb}
\makeatletter
  \newcommand\figcaption{\def\@captype{figure}\caption}
  \newcommand\tabcaption{\def\@captype{table}\caption}
\makeatother
\begin{document}

\title{Towards Faster Training of Global Covariance Pooling Networks by Iterative Matrix Square Root Normalization} 
\author{Peihua Li, Jiangtao Xie, Qilong Wang, Zilin Gao\\
	Dalian University of Technology\\
	{\tt\small peihuali@dlut.edu.cn}
}

\maketitle
\thispagestyle{empty}

\begin{abstract}
Global covariance pooling in convolutional neural networks  has achieved impressive improvement over the classical first-order pooling. Recent works have shown matrix square root normalization plays a central role in achieving state-of-the-art performance. However, existing methods depend heavily  on eigendecomposition (EIG)  or singular value decomposition (SVD), suffering from inefficient training due to limited support of EIG and SVD on  GPU. Towards addressing this problem, we propose an iterative matrix square root normalization method for fast end-to-end  training of global covariance pooling networks. At the core of our method is  a meta-layer  designed with loop-embedded directed graph structure. The meta-layer  consists of three consecutive nonlinear structured layers, which perform pre-normalization, coupled matrix  iteration and post-compensation, respectively. Our method is much faster than EIG or SVD based ones, since it involves only matrix multiplications, suitable for parallel implementation on GPU.  Moreover, the proposed network with ResNet architecture can converge in much less epochs, further accelerating network training. On large-scale ImageNet, we achieve competitive performance superior to existing counterparts. By  finetuning our models pre-trained on ImageNet, we establish  state-of-the-art results on three challenging fine-grained benchmarks. The  source code and network models will be available at  \href{http://www.peihuali.org/iSQRT-COV}{http://www.peihuali.org/iSQRT-COV}. 
\end{abstract}

\section{Introduction}

Deep convolutional neural networks (ConvNets) have made significant progress in the past years, achieving recognition accuracy surpassing human beings in large-scale object recognition~\cite{He_2015_ICCV}.{ \let\thefootnote\relax\footnote{The work was supported by National Natural Science Foundation of China (No. 61471082). Peihua Li is the corresponding author.}}The ConvNet models pre-trained on ImageNet~\cite{imagenet_cvpr09} have been proven to benefit a multitude of other computer vision tasks, ranging from fine-grained visual categorization (FGVC)~\cite{lin2015bilinear}, object detection~\cite{Redmon_2017_CVPR}, semantic segmentation~\cite{Long_2015_CVPR} to scene parsing~\cite{zhou2017scene},   where labeled data are insufficient for training from scratch. The common layers such as convolution, non-linear rectification, pooling and batch normalization~\cite{DBLP:journals/corr/IoffeS15} have  become off-the-shelf commodities, widely supported on devices including workstations, PCs and embedded systems.

Although the architecture of ConvNet has greatly evolved in the past years, its basic layers  largely keep unchanged~\cite{LeNet1989,Krizhevsky2012ImageNet}.  Recently, researchers have shown increasing interests in exploring structured layers to  enhance representation capability of networks~\cite{Ionescu_2015_ICCV,lin2015bilinear,Arandjelovic_2016_CVPR,LiYunSheng_2017_ICCV}. One particular kind of structured layer is concerned with global covariance pooling after the last convolution layer, which has shown impressive improvement over the classical first-order pooling, successfully used in FGVC~\cite{lin2015bilinear},  visual question answering~\cite{Kafle_2017_ICCV} and  video action recognition~\cite{WangYunbo_2017_CVPR}.   Very recent works  have demonstrated that  matrix square root normalization of global covariance pooling plays a key role in achieving state-of-the-art performance in both large-scale visual recognition~\cite{Li_2017_ICCV} and  challenging FGVC~\cite{lin2017improved,Wang_2017_CVPR}.

\begin{table*}[htb]
\setlength\tabcolsep{4pt}
\renewcommand{\baselinestretch}{1.05}
\footnotesize
\begin{center}
\begin{tabular}{|l|c|c|c|c|c|}
\hline
Method                  & Forward  Prop. (FP)          & Backward Prop. (BP)  &  \parbox{0.6in}{\centering  \vspace{1pt} CUDA\\ support \vspace{1pt}} & \parbox{0.7in}{\centering Scalability to \\ multi-GPUs} & \parbox{0.9in}{Large-scale (LS) or\\ Small-scale (SS)} \\
\hline
\hline
MPN-COV~\cite{Li_2017_ICCV}                  &  \parbox{0.9in}{\centering  \textbf{\color{red}EIG algorithm}}                   &   \parbox{1.2in}{\centering  BP of EIG}             &   \parbox{0.5in}{\centering limited} & limited & LS only \\
\hline
\parbox{0.6in}{G$^2$DeNet~\cite{Wang_2017_CVPR}}          &  \parbox{0.9in}{\centering  \textbf{\color{red}SVD algorithm}}   & \parbox{1.2in}{\centering BP of SVD}          &  \parbox{0.6in}{\centering limited} & limited & SS only  \\
\hline
\multirow{2}{*}{\parbox{1.0in}{Improved  B-CNN~\cite{lin2017improved}}}          &  \parbox{0.9in}{\centering Newton-Schulz Iter.}   & \parbox{1.7in}{\centering \vspace{2pt} \textbf{\color{red}BP by Lyapunov equation \\ (SCHUR or EIG required)}\vspace{1pt}}          &  \multirow{2}{*}{limited} & \multirow{2}{*}{limited} & \multirow{2}{*}{SS only}  \\
\cline{2-3}
& \textbf{\color{red}SVD algorithm} & BP of SVD & & & \\
\hline
\parbox{0.9in}{\vspace{4pt}iSQRT-COV (ours)\vspace{3pt}} &  \parbox{0.9in}{\centering Newton-Schulz Iter.}  & \parbox{1.2in}{\centering  BP of Newton-Schulz Iter.}         &  good  & good & LS+SS  \\
\hline
\end{tabular}
\end{center}
\renewcommand{\baselinestretch}{1.0}
\caption{Differences between our iSQRT-COV and related methods. The bottleneck operations are marked with red, bold text.}
\label{table:summary-differences}
\end{table*}

For computing matrix square root, existing methods depend heavily on eigendecomposition (EIG) or singular value decomposition (SVD)~\cite{Li_2017_ICCV,Wang_2017_CVPR,lin2017improved}. However, fast implementation  of EIG or SVD on GPU  is an open problem, which is limitedly supported on NVIDIA CUDA platform, significantly slower than their CPU counterparts~\cite{Ionescu_2015_ICCV,lin2017improved}. As such,  existing methods opt for EIG or SVD on CPU for computing matrix square root. Nevertheless, current implementations of meta-layers depending on CPU are far from ideal, particularly for multi-GPU configuration. Since GPUs with powerful parallel computing ability have to be interrupted and await  CPUs with limited parallel ability, their  concurrency and throughput are greatly restricted.

In~\cite{lin2017improved}, for the purpose of fast forward propagation (FP), Lin and Maji  use  Newton-Schulz iteration (called modified Denman-Beavers iteration therein) algorithm, which is 
proposed in~\cite{Higham:2008:FM}, to compute matrix square-root.  Unfortunately, for backward propagation (BP), they compute the gradient  through  Lyapunov equation solution which depends on the GPU unfriendly Schur-decomposition (SCHUR) or EIG. Hence, the training in~\cite{lin2017improved}~is expensive though FP which involves only matrix multiplication runs very fast. 
Inspired  by that work, we propose a fast  end-to-end training method, called iterative  matrix square root normalization of covariance pooling (iSQRT-COV), depending on  Newton-Schulz iteration in both forward and backward propagations. 

At the core of iSQRT-COV is a meta-layer  with loop-embedded directed graph  structure, specifically designed  for  ensuring both  convergence of Newton-Schulz iteration and  performance of global covariance pooling networks. The meta-layer consists of three consecutive  structured layers,  performing pre-normalization, coupled matrix iteration and post-compensation, respectively. We derive the gradients associated with the involved non-linear layers based on matrix backpropagation theory~\cite{Ionescu_2015_ICCV}. The design of sandwiching   Newton-Schulz iteration using  pre-normalization by Frobenius norm or trace and post-compensation is essential, which, as far as we know, did not appear in previous literature (e.g. in~\cite{Higham:2008:FM} or~\cite{lin2017improved} ). The pre-normalization guarantees  convergence of Newton-Schulz (NS) iteration, while post-compensation plays a key role in achieving state-of-the-art  performance  with prevalent deep ConvNet architectures, e.g. ResNet~\cite{He_2016_CVPR}. 
The main differences between our method and other  related works\footnote{It is worth noting that, after CVPR submission deadline,  authors of~\cite{lin2017improved} release  code of improved B-CNN together with a scheme similar to ours, in which  BP of Newton-Schulz iteration is implemented using Autograd package  in PyTorch. We note  that (1) that scheme is parallel to our work, and (2) they only provide pieces of code but do not train using BP of Newton-Schulz iteration  on any real-world benchmarks. } are summarized in Tab.~\ref{table:summary-differences}.

\section{Related Work}\label{section:related-work}

B-CNN is one of the first end-to-end covariance  pooling ConvNets~\cite{lin2015bilinear,Ionescu_2015_ICCV}. It performs element-wise  square root normalization followed by $\ell_{2}-$normalization for covariance matrix, achieving impressive performance in FGVC task. Improved B-CNN~\cite{lin2017improved} shows that additional matrix square root normalization before element-wise square root and $\ell_{2}-$normalization  can further attain  large  improvement. In training process, they perform FP using Newton-Schulz iteration  or using SVD, and perform BP by solving Lyapunov equation or compute gradients associated with SVD. In any  case, improved B-CNN suffers from GPU unfriendly  SVD, SCHUR or EIG and so network training is expensive. Our iSQRT-COV differs from~\cite{lin2017improved} in three aspects. First, both FP and BP of our method are based on Newton-Schulz iteration, making network training very efficient as only GPU friendly matrix multiplications are involved. Second, we propose sandwiching    Newton-Schulz iteration using  pre-normalization  and post-compensation which is essential and plays a key role in training extremely  deep ConvNets. Finally, we evaluate extensively on both large-scale ImageNet and  on three popular fine-grained benchmarks. 

\begin{figure*}[t]
\begin{center}
   \includegraphics[width=1.0\linewidth]{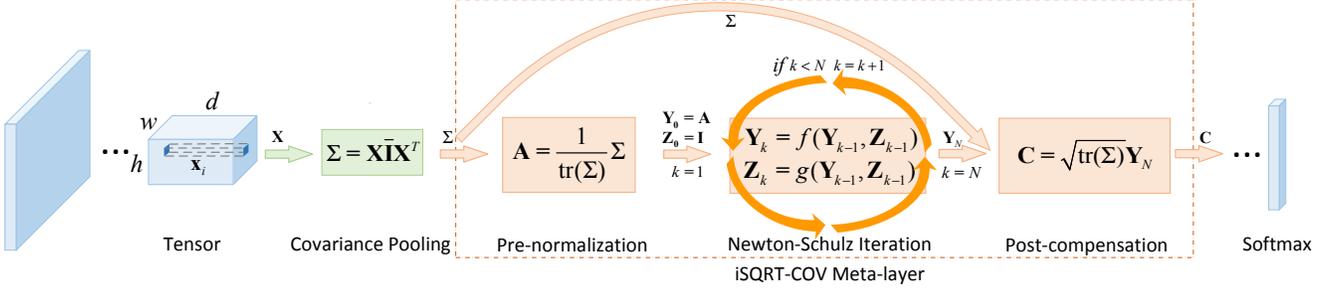}
\end{center}
\caption{Proposed iterative matrix square root normalization of covariance pooling  (iSQRT-COV) network. After the last convolution layer, we perform second-order pooling by estimating a covariance matrix. We design a meta-layer with loop-embedded directed graph structure for computing approximate square root of covariance matrix. The meta-layer  consists of  three  nonlinear structured layers, performing pre-normalization, coupled  Newton-Schulz iteration and post-compensation, respectively.  See Sec. ~\ref{section:proposed-method} for  notations and  details.}
\label{fig:overview}
\end{figure*}

In~\cite{Li_2017_ICCV}, matrix power normalized covariance pooling method (MPN-COV) is proposed  for large-scale visual recognition. It achieves impressive  improvements over first-order pooling with  AlexNet~\cite{Krizhevsky2012ImageNet}, VGG-Net~\cite{DBLP:conf/bmvc/ChatfieldSVZ14,Simonyan15} and ResNet~\cite{He_2016_CVPR} architectures. MPN-COV has shown that,  given a small number of high-dimensional features,  matrix power is consistent with shrinkage principle of robust covariance estimation, and matrix square root can be derived as a robust covariance estimator via  a von Neumann regularized maximum likelihood estimation~\cite{Wang_2016_CVPR}. It is also shown that matrix power normalization approximately yet effectively exploits geometry of the manifold of covariance matrices, superior to matrix logarithm normalization~\cite{Ionescu_2015_ICCV} for high-dimensional  features. All computations of MPN-COV meta-layer are implemented with NVIDIA cuBLAS library running on GPU, except EIG which runs on CPU.

G$^2$DeNet~\cite{Wang_2017_CVPR} is concerned with inserting global Gaussian distributions into  ConvNets for end-to-end learning. In G$^2$DeNet, each Gaussian  is identified as square root of a symmetric positive definite matrix based on Lie group structure of Gaussian manifold~\cite{LE2MG}. The matrix square root  plays a central role in obtaining the competitive performance~\cite[Tab. 1 \& Tab. 5]{Wang_2017_CVPR}.   Compact bilinear pooling (CBP)~\cite{Gao_2016_CVPR} clarifies that bilinear pooling is closely related to the second-order polynomial kernel, and presents two compact representations via low-dimensional feature maps for kernel approximation. Kernel pooling~\cite{Cui_2017_CVPR} approximates Gaussian RBF kernel to a given order through compact explicit feature maps, aiming to characterize higher order feature interactions.  Cai et al.~\cite{Cai_2017_ICCV} introduce a polynomial kernel based predictor to model higher-order statistics of convolutional features across multiple layers.

\section{Proposed iSQRT-COV Network}\label{section:proposed-method}

In this section, we first give an overview of the proposed iSQRT-COV network. Then we describe matrix square root computation and its forward  propagation. We finally  derive the corresponding backward gradients.

\subsection{Overview of Method}

The flowchart of the proposed network is shown in Fig.~\ref{fig:overview}. Let  output of the last convolutional layer (with ReLU) be a $h\times w\times d$ tensor  with spatial height $h$, width $w$ and channel $d$. We reshape the tensor to a feature matrix $\mathbf{X}$ consisting of $n=wh$ features of $d-$dimension. Then we perform second-order pooling by computing the covariance matrix $\boldsymbol{\Sigma}=\mathbf{X}\bar{\mathbf{I}}\mathbf{X}^{T}$, where $\bar{\mathbf{I}}=\frac{1}{n}(\mathbf{I}-\frac{1}{n}\mathbf{1})$,  $\mathbf{I}$ and $\mathbf{1}$ are the $n\times n$ identity matrix and matrix of all ones, respectively.

Our meta-layer is designed to have loop-embedded directed graph structure, consisting of three consecutive  nonlinear structured layers. The purpose of the first layer (i.e., pre-normalization) is to guarantee the convergence of the following  Newton-Schulz iteration, achieved by dividing the covariance matrix by its trace (or Frobenius norm).  The second layer is of loop structure, repeating the coupled  matrix equations involved in Newton-Schulz iteration a fixed number of times, for computing  approximate matrix  square root. The pre-normalization nontrivially changes  data magnitudes, so we  design the third layer (i.e., post-compensation) to counteract the adverse effect by multiplying trace (or Frobenius norm) of the square root of the covariance matrix.
As the output of our meta-layer  is a symmetric matrix, we concatenate its upper triangular entries forming an $d(d+1)/2$-dimensional vector, submitted to the subsequent layer of the ConvNet.

\subsection{Matrix Square Root and Forward Propagation}\label{subsection:forward}

Square roots of matrices, particularly covariance matrices which are symmetric  positive (semi)definite (SPD), find applications in a variety of fields including computer vision, medical imaging~\cite{doi:10.1080/02664763.2015.1080671} and chemical physics~\cite{2007JChPh.126l4104J}.  It is well-known  any SPD matrix has a unique square root which can be computed accurately  by EIG or SVD. Briefly, let $\mathbf{A}$ be an SPD matrix and it has EIG $\mathbf{A}=\mathbf{U}\mathrm{diag}(\lambda_{i})\mathbf{U}^{T}$, where $\mathbf{U}$ is orthogonal and $\mathrm{diag}(\lambda_{i})$ is a diagonal matrix of eigenvalues $\lambda_{i}$ of $\mathbf{A}$. Then $\mathbf{A}$ has a  square root $\mathbf{Y}=\mathbf{U}\mathrm{diag}(\lambda_{i}^{1/2})\mathbf{U}^{T}$, i.e., $\mathbf{Y}^{2}=\mathbf{A}$. Unfortunately, both EIG and SVD are not well supported on GPU.

\vspace{-8pt}\paragraph{Newton-Schulz Iteration}
Higham~\cite{Higham:2008:FM} studied a class of methods for iteratively computing  matrix square root. These methods, termed as  Newton-Pad\'{e} iterations, are developed based on the connection between  matrix sign function and matrix square root, together with  rational Pad\'{e} approximation.  Specifically, for computing the square root $\mathbf{Y}$ of $\mathbf{A}$,   given $\mathbf{Y}_{0}=\mathbf{A}$ and $\mathbf{Z}_{0}=\mathbf{I}$, for $k=1,\cdots, N$, the coupled iteration  takes the following form~\cite[Chap. 6.7]{Higham:2008:FM}:
\begin{align}\label{equ:coupled-equation0}
\mathbf{Y}_{k}&=\mathbf{Y}_{k-1}p_{lm}(\mathbf{Z}_{k-1}\mathbf{Y}_{k-1})q_{lm}(\mathbf{Z}_{k-1}\mathbf{Y}_{k-1})^{-1}\nonumber \\
\mathbf{Z}_{k}&=p_{lm}(\mathbf{Z}_{k-1}\mathbf{Y}_{k-1})q_{lm}(\mathbf{Z}_{k-1}\mathbf{Y}_{k-1})^{-1}\mathbf{Z}_{k-1},
\end{align}
where $p_{lm}$ and $q_{lm}$ are polynomials, and $l$ and $m$ are non-negative integers. Eqn.~(\ref{equ:coupled-equation0}) converges only locally:  if $\|\mathbf{A}-\mathbf{I}\|<1$ where  $\|\cdot\|$  denotes any induced (or consistent) matrix norm, $\mathbf{Y}_{k}$ and $\mathbf{Z}_{k}$ quadratically converge to $\mathbf{Y}$ and $\mathbf{Y}^{-1}$, respectively. The family of coupled iteration is  stable in that small errors in the previous iteration will not be amplified.  The case of $l=0, m=1$ called \textit{Newton-Schulz iteration}   fits for our purpose as no GPU unfriendly matrix inverse is involved:
\begin{align}\label{equ:coupled-equation}
\mathbf{Y}_{k}&=\dfrac{1}{2}\mathbf{Y}_{k-1}(3\mathbf{I}-\mathbf{Z}_{k-1}\mathbf{Y}_{k-1})\nonumber \\
\mathbf{Z}_{k}&=\dfrac{1}{2}(3\mathbf{I}-\mathbf{Z}_{k-1}\mathbf{Y}_{k-1})\mathbf{Z}_{k-1}. 
\end{align}
Clearly Eqn. (\ref{equ:coupled-equation}) involves only matrix product, suitable for parallel implementation on GPU. Compared to \textit{accurate} square root  computed by EIG, one can only obtain \textit{approximate} solution with a small number of iterations. We determine the  number of iterations $N$  by cross-validation. Interestingly, compared to EIG or SVD based methods, experiments  on large-scale ImageNet show that  we can obtain matching or marginally better performance under AlexNet architecture~(Sec.~\ref{section:ImageNet-AlexNet}) and  better performance under ResNet architecture~(Sec.~\ref{section:ImageNet-ResNet}), using no more than 5 iterations.

\vspace{-8pt}\paragraph{Pre-normalization and Post-compensation} As Newton-Schulz iteration only converges locally, we pre-normalize $\boldsymbol{\Sigma}$ by trace or Frobenius norm, i.e.,
\begin{align}\label{equ:pre-normalization}
\mathbf{A}=\frac{1}{\mathrm{tr}(\boldsymbol{\Sigma})}\boldsymbol{\Sigma} \;\; \text{or}\;\;
\frac{1}{\|\boldsymbol{\Sigma}\|_{F}}\boldsymbol{\Sigma}.
\end{align}
Let $\lambda_{i}$ be eigenvalues of $\boldsymbol{\Sigma}$, arranged in nondecreasing order. As $\mathrm{tr}(\boldsymbol{\Sigma})=\sum_{i}\lambda_{i}$ and $\|\boldsymbol{\Sigma}\|_{F}=\sqrt{\sum_{i}\lambda_{i}^{2}}$, it is easy to see that $\|\boldsymbol{\Sigma}-\mathbf{I}\|_{2}$, which equals to the largest singular value of $\boldsymbol{\Sigma}-\mathbf{I}$, is $1-\frac{\lambda_{1}}{\sum_{i}\lambda_{i}}$ and $1-\frac{\lambda_{1}}{\sqrt{\sum_{i}\lambda_{i}^{2}}}$ for the case of trace and Frobenius norm, respectively, both  less than 1. Hence, the convergence condition is satisfied. 

The above pre-normalization of  covariance matrix nontrivially changes the data magnitudes such that it produces adverse effect on network. Hence, to counteract this change, after the Newton-Schulz iteration,  we accordingly perform post-compensation, i.e.,
\begin{align}\label{equ:post-compensation}
\mathbf{C}=\sqrt{\mathrm{tr}(\boldsymbol{\Sigma)}}\mathbf{Y}_{N}\;\; \text{or}\;\; \mathbf{C}=\sqrt{\|\boldsymbol{\Sigma}\|_{F}}\mathbf{Y}_{N}.
\end{align}
An alternative scheme to  counterbalance the influence incurred by  pre-normalization is Batch Normalization (BN)~\cite{DBLP:journals/corr/IoffeS15}. One may even consider without using any post-compensation. However, our experiment on ImageNet  has shown that, without post-normalization, prevalent ResNet~\cite{He_2016_CVPR} fails to converge, while our scheme outperforms BN by about 1\% (see~\ref{section:ImageNet-ResNet} for details).

\subsection{Backward Propagation (BP)}\label{subsection:backward}

The gradients associated with the structured layers are derived using matrix backpropagation methodology~\cite{IonescuVS15}, which establishes the chain rule of a general matrix function by  first-order Taylor approximation.  Below we take \textit{pre-normalization by trace} as an example, deriving the corresponding gradients.

\paragraph{BP of Post-compensation} Given $\frac{\partial l}{\partial \mathbf{C}}$ where $l$ is the loss function, the chain rule is of the form 
$
\mathrm{tr}\big(\big(\frac{\partial l}{\partial \mathbf{C}}\big)^{T}\mathrm{d}\mathbf{C}\big)=\mathrm{tr}\big(\big(\frac{\partial l}{\partial \mathbf{Y}_{N}}\big)^{T}\mathrm{d}\mathbf{Y}_{N}+\big(\frac{\partial l}{\partial \boldsymbol{\Sigma}}\big)^{T}\mathrm{d}\boldsymbol{\Sigma}\big), 
$
where $\mathrm{d}\mathbf{C}$ denotes variation of $\mathbf{C}$. After some manipulations, we have
\begin{align}\label{equ:BP-post-trace}
\dfrac{\partial l}{\partial \mathbf{Y}_{N}}&=\sqrt{\mathrm{tr}(\boldsymbol{\Sigma)}}\dfrac{\partial l}{\partial \mathbf{C}}  \nonumber \\
\dfrac{\partial l}{\partial \boldsymbol{\Sigma}}\Big|_{\mathrm{post}}&=\dfrac{1}{2\sqrt{\mathrm{tr}(\boldsymbol{\Sigma)}}}\mathrm{tr}\Big(\Big(\dfrac{\partial l}{\partial \mathbf{C}}\Big)^{T}\mathbf{Y}_{N}\Big)\mathbf{I}.
\end{align}

\paragraph{BP of Newton-Schulz Iteration} Then we are to compute the partial derivatives of the loss function with respect to $\frac{\partial l}{\partial \mathbf{Y}_{k}}$ and $\frac{\partial l}{\partial \mathbf{Z}_{k}}$, $k=N-1, \ldots, 1$, given $\frac{\partial l}{\partial \mathbf{Y}_{N}}$ computed by Eqn.~(\ref{equ:BP-post-trace}) and  $\frac{\partial l}{\partial \mathbf{Z}_{N}}=0$. As the covariance matrix $\boldsymbol{\Sigma}$ is symmetric, it is easy to see from Eqn.~(\ref{equ:coupled-equation}) that $\mathbf{Y}_{k}$ and $\mathbf{Z}_{k}$ are both symmetric. According to the chain rules (omitted hereafter  for simplicity) of matrix backpropagation and after some manipulations, $k=N, \ldots, 2$, we can derive
\begin{align}\label{equ:BP-coupled-equations}
\dfrac{\partial l}{\partial \mathbf{Y}_{k-1}}=&\dfrac{1}{2}\Big(\dfrac{\partial l}{\partial \mathbf{Y}_{k}}\Big(3\mathbf{I}-\mathbf{Y}_{k-1}\mathbf{Z}_{k-1}\Big)-\mathbf{Z}_{k-1}\dfrac{\partial l}{\partial \mathbf{Z}_{k}}\mathbf{Z}_{k-1}\nonumber \\
&-\mathbf{Z}_{k-1}\mathbf{Y}_{k-1}\dfrac{\partial l}{\partial \mathbf{Y}_{k}}\Big)\nonumber\\
\dfrac{\partial l}{\partial \mathbf{Z}_{k-1}}=&\dfrac{1}{2}\Big(\Big(3\mathbf{I}-\mathbf{Y}_{k-1}\mathbf{Z}_{k-1}\Big)\dfrac{\partial l}{\partial \mathbf{Z}_{k}}-\mathbf{Y}_{k-1}\dfrac{\partial l}{\partial \mathbf{Y}_{k}}\mathbf{Y}_{k-1}\nonumber\\
&-\dfrac{\partial l}{\partial \mathbf{Z}_{k}}\mathbf{Z}_{k-1}\mathbf{Y}_{k-1}\Big).
\end{align}
The final step of this layer is concerned with the partial derivative with respect to $\frac{\partial l}{\partial \mathbf{A}}$, which is given by
\begin{align}\label{equ:BP-gradient-A}
\dfrac{\partial l}{\partial \mathbf{A}}=\dfrac{1}{2}\Big(\dfrac{\partial l}{\partial \mathbf{Y}_{1}}\Big(3\mathbf{I}-\mathbf{A}\Big)-\dfrac{\partial l}{\partial \mathbf{Z}_{1}}-\mathbf{A}\dfrac{\partial l}{\partial \mathbf{Y}_{1}}\Big).
\end{align}

\paragraph{BP of Pre-normalization} Note that here we need to combine the gradient of the loss function $l$ with respect to $\boldsymbol{\Sigma}$, backpropagated from the post-compensation layer. As such, by referring to Eqn.~(\ref{equ:pre-normalization}), we make similar derivations as before and obtain 
\begin{align}\label{equ:BP-pre-trace}
\dfrac{\partial l}{\partial \boldsymbol{\Sigma}}=&-\dfrac{1}{({\mathrm{tr}(\boldsymbol{\Sigma)}})^2}\mathrm{tr}\Big(\Big(\dfrac{\partial l}{\partial \mathbf{A}}\Big)^{T}\boldsymbol{\Sigma}\Big)\mathbf{I}+\dfrac{1}{\mathrm{tr}(\boldsymbol{\boldsymbol{\Sigma}})}\dfrac{\partial l}{\partial \mathbf{A}}\nonumber\\
&+\dfrac{\partial l}{\partial \boldsymbol{\Sigma}}\Big|_{\mathrm{post}}.
\end{align}

If we adopt \textit{pre-normalization by Frobenius norm}, the gradients associated with post-compensation  become
\begin{align}\label{equ:BP-post-fro}
\dfrac{\partial l}{\partial \mathbf{Y}_{N}}&=\sqrt{\|\boldsymbol{\Sigma}\|_{F}}\dfrac{\partial l}{\partial \mathbf{C}} \nonumber \\
\dfrac{\partial l}{\partial \boldsymbol{\Sigma}}\Big|_{\mathrm{post}}&=\dfrac{1}{2\|\boldsymbol{\Sigma}\|_{F}^{3/2}}\mathrm{tr}\Big(\Big(\dfrac{\partial l}{\partial \mathbf{C}}\Big)^{T}\mathbf{Y}_{N}\Big)\boldsymbol{\Sigma},
\end{align}
and that with respect to  pre-normalization  is
\begin{align}\label{equ:BP-pre-trace}
\dfrac{\partial l}{\partial \boldsymbol{\Sigma}}=&-\dfrac{1}{\|\boldsymbol{\Sigma}\|_{F}^{3}}\mathrm{tr}\Big(\Big(\dfrac{\partial l}{\partial \mathbf{A}}\Big)^{T}\boldsymbol{\Sigma}\Big)\boldsymbol{\Sigma}+\dfrac{1}{\|\boldsymbol{\Sigma}\|_{F}}\dfrac{\partial l}{\partial \mathbf{A}}\nonumber\\
&+\dfrac{\partial l}{\partial \boldsymbol{\Sigma}}\Big|_{\mathrm{post}},
\end{align}
while the backward gradients of Newton-Schulz iteration (\ref{equ:BP-coupled-equations}) keep unchanged.

Finally, given $\frac{\partial l}{\partial \boldsymbol{\Sigma}}$, one can derive the gradient of the loss function $l$ with respect to  input matrix $\mathbf{X}$, which takes the following form~\cite{Li_2017_ICCV}:
\begin{align}\label{equ:compute-cov-backward}
\dfrac{\partial l}{\partial \mathbf{X}}=\bar{\mathbf{I}}\mathbf{X}\bigg(\dfrac{\partial l}{\partial \boldsymbol{\Sigma}}+\bigg(\dfrac{\partial l}{\partial \boldsymbol{\Sigma}}\bigg)^{T}\bigg).
\end{align}

\section{Experiments}\label{section:experiments}

We  evaluate the proposed method on both large-scale image classification and challenging fine-grained visual categorization  tasks. We make experiments using two PCs each of which is equipped with a 4-core Intel i7-4790k@4.0GHz CPU, 32G RAM, 512GB Samsung  PRO SSD and two Titan Xp GPUs.  We implement our networks using  MatConvNet~\cite{vedaldi15matconvnet} and Matlab2015b, under Ubuntu 14.04.5 LTS.

\subsection{Datasets and Our Meta-layer Implementation}
\vspace{4pt}\noindent\textbf{Datasets}\quad For large-scale image classification, we adopt \textit{ImageNet LSVRC2012 dataset}~\cite{imagenet_cvpr09} with 1,000 object categories.  The dataset contains  1.28M images for training, 50K images for validation  and 100K images for testing (without published labels). As in~\cite{DBLP:journals/corr/IoffeS15,He_2016_CVPR}, we report the results on the validation set. For fine-grained categorization, we use three popular \textit{fine-grained benchmarks}, i.e.,  CUB-200-2011(Birds)~\cite{Wah2011The}, FGVC-aircraft (Aircrafts)~\cite{Maji2013Fine} and Stanford cars (Cars)~\cite{Krause20133D}.  The Birds dataset contains 11,788 images from 200 species, with large intra-class variation but small inter-class variation. The Aircrafts dataset includes 100 aircraft classes and a total of 10,000 images with small background noise but higher inter-class similarity. The Cars dataset  consists of 16,185 images from 196 classes. For all datasets, we adopt the provided training/test split, using neither bounding boxes nor part annotations.

\vspace{4pt}\noindent\textbf{Implementation of iSQRT-COV Meta-layer}\quad
We encapsulate our code in three computational blocks, which implement  forward\&backward  computation of pre-normalization layer, Newton-Schulz iteration layer and post-compensation layer, respectively. The code is written in C++ based on NVIDIA \href{http://docs.nvidia.com/cuda/cublas/}{cuBLAS} on top of  CUDA toolkit 8.0.  In addition, we write code in C++ based on cuBLAS for computing covariance matrices. We create MEX files so that the above subroutines can be called  in Matlab environment. For AlexNet, we insert  our meta-layer after the last convolution layer (with ReLU), which outputs an $13\times 13\times  256$ tensor. For ResNet architecture, as suggested~\cite{Li_2017_ICCV}, we do not perform downsampling for the last set of convolutional blocks, and  add one $1\times 1$ convolution with $d=256$ channels after the last sum layer (with ReLU). The added $1\times 1$ convolution layer outputs an $14\times 14\times 256$ tensor. Hence, with both architectures, the covariance matrix $\boldsymbol{\Sigma}$ is of size $256\times 256$ and our meta-layer outputs an $d(d+1)/2\approx32\mathrm{K}$-dimensional vector as the image representation.

\subsection{Evaluation with AlexNet on ImageNet}\label{section:ImageNet-AlexNet}

In the first part of experiments, we analyze, with AlexNet architecture, the design choices of our iSQRT-COV method, including the number of Newton-Schulz iterations,  time and memory usage, and behaviors of different pre-normalization methods.  We select AlexNet because it runs faster with shallower depth, and the results can  extrapolate to deeper networks  which mostly follow its architecture design.

\begin{figure}
\centering
\begin{minipage}[b]{0.65\linewidth}
\centering
\includegraphics[width=1.0\textwidth]{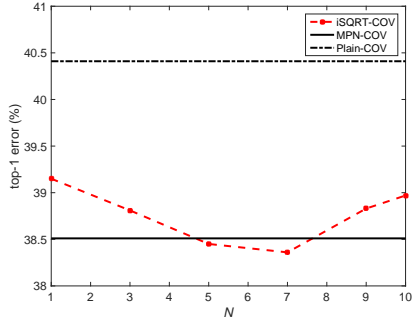}
\end{minipage}
\caption{Impact of  number $N$ of Newton-Schulz iterations on iSQRT-COV with AlexNet architecture on ImageNet.}
\label{fig:impact_iteration-N}
\end{figure}

We follow~\cite{Li_2017_ICCV} for color augmentation and weight initialization, adopting BN and no dropout. We use SGD with  a mini-batch of 128,  unless otherwise stated. The momentum is  0.9 and  weight decay is 0.0005. We train iSQRT-COV networks from scratch in 20 epochs where  learning rate follows exponential decay  $10^{-1.1} \to 10^{-5}$. All training and test images are uniformly resized with shorter sides of 256. During training we randomly crop a $224\times224$ patch from each image or its horizontal flip. We make inference on one single  $224\times 224$ center crop from a test image. 

\vspace{4pt}\noindent\textbf{Impact of  Number $N$ of Newton-Schulz Iterations}\quad 
Fig.~\ref{fig:impact_iteration-N} shows top-1 error rate as a function of  number  of Newton-Schulz iterations in Eqn.  (\ref{equ:coupled-equation}). Plain-COV indicates simple covariance pooling without any normalization. With one single iteration, our method outperforms Plain-COV by $1.3\%$. As iteration number grows, the error rate of iSQRT-COV gradually declines. With 3 iterations, iSQRT-COV is comparable to MPN-COV, having only 0.3\% higher error rate, while performing marginally better than MPN-COV between 5 and 7 iterations. After $N=7$, the error rate consistently increases, indicating growth of iteration number is not helpful for improving accuracy. As larger $N$ incurs higher computational cost, to balance efficiency and accuracy, we set $N$ to 5  in the remaining experiments. Notably, the approximate square root normalization improves a little over the accurate one obtained via EIG. This interesting problem will be discussed in  Sec.~\ref{section:ImageNet-ResNet}, where iSQRT-COV is further evaluated on substantially deeper ResNets.

\begin{table}[t]
	\setlength\tabcolsep{1.0pt}
	\subtable[Time of FP+BP (ms) taken and  memory (MB) used by single meta-layer. Numbers in parentheses indicate FP time.]{\label{table:single-layer}
		\renewcommand{\baselinestretch}{1.1}
		\begin{minipage}[t]{1.0\linewidth}
			\centering
			\footnotesize
			\begin{tabular}{|l|l|c|c|c|c|}
				\hline
				\multicolumn{2}{|c|}{Method}                               &  Language    &  bottleneck  & Time & Memory \\
				\hline
				\hline
				\multicolumn{2}{|c|}{iSQRT-COV ($N$$=$$3$)}                & \multirow{2}{*}{ \parbox{0.20in}{C++}}& \multirow{2}{*}{ \parbox{0.2in}{N/A}}    & \color{blue}{\textbf{0.81}} (\color{blue}{\textbf{0.26}}) & 0.627   \\
				\multicolumn{2}{|c|}{iSQRT-COV ($N$$=$$5$)}                &          &   & \textbf{1.41} (\textbf{0.41})   & \textbf{1.129}  \\
				\hline
				\multicolumn{2}{|c|}{MPN-COV \cite{Li_2017_ICCV}}          &  C++\&M & EIG & 2.58 (2.41)   & 0.377 \\
				\hline
				\multirow{4}{*}{\parbox{0.35in}{Impro. \\B-CNN \\\cite{lin2017improved}}} & FP and BP based  & \multirow{4}{*}{M} & \multirow{4}{*}{\parbox{0.20in}{SVD or EIG}} & \multirow{2}{*}{13.51 (11.19)} & \multirow{4}{*}{0.501}\\
				& on SVD   & & &  & \\
				\cline{2-2}\cline{5-5}
				&  FP by NS Iter.,    & & & \multirow{2}{*}{13.91 (2.09)} & \\
				&  BP by Lyap.   & & & & \\
				\hline
				\multicolumn{2}{|c|}{G$^2$DeNet \cite{Wang_2017_CVPR}}     &  M      & SVD  & 8.56 (4.76)  & 0.505   \\
				\hline
			\end{tabular}
	\end{minipage}}
	
	\subtable[Time (ms) taken by  matrix decomposition (single precision arithmetic)]{\label{table:time-matrix-decomposition}
		\renewcommand{\baselinestretch}{1.05}
		\begin{minipage}[t]{1.0\linewidth}
			\centering
			\footnotesize
			\begin{tabular}{|c|c|c|c|}
				\hline
				Algorithm                            & \parbox{0.50in}{\centering CUDA\\cuSOLVER}    & \parbox{0.7in}{\centering \vspace{2pt} Matlab\\(CPU function)\vspace{2pt}} & \parbox{0.70in}{\centering Matlab\\(GPU function)} \\
				\hline
				\hline
				EIG                                  &    21.3 & 1.8    & 9.8     \\
				SVD                                  &    52.2 & 4.1    & 11.9     \\
				\hline
			\end{tabular}
	\end{minipage}}
	\caption{Comparison of time and memory usage with AlexNet architecture. The size of covariance matrix is $256\times 256$.}
	\label{table:time-comparison}
\end{table}

\begin{figure}[thb]
\vspace{2pt}
\setlength\tabcolsep{8pt}
\renewcommand{\baselinestretch}{1.0}
\footnotesize
\centering
\begin{minipage}[b]{0.65\linewidth}
\centering
\includegraphics[width=1.0\textwidth]{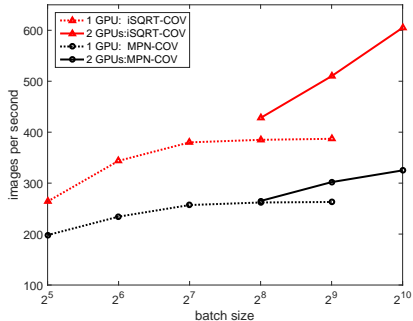}
\end{minipage}
\caption{Images per second (FP$+$BP) of network training with AlexNet architecture.}
\label{fig:Comparison_time}
\end{figure}

\vspace{4pt}\noindent\textbf{Time and Memory Analysis}\quad We  compare  time and memory consumed by  single meta-layer of different methods. We use  public code for \href{https://github.com/jiangtaoxie/MPN-COV-ConvNet}{MPN-COV},  \href{http://www.peihuali.org/publications/G2DeNet/G2DeNet-FGVC-v1.0.zip}{G$^{2}$DeNet} and \href{https://bitbucket.org/tsungyu/bcnn}{improved B-CNN} released by the respective authors. As shown in Tab.~\ref{table:time-comparison}\subref{table:single-layer}, iSQRT-COV~($N=3$) and iSQRT-COV~($N=5$) are 3.1x faster and 1.8x faster than MPN-COV, respectively. Furthermore, iSQRT-COV~($N=5$) is  five  times more efficient than  improved B-CNN and G$^2$DeNet. For improved B-CNN, the forward computation of Newton-Schulz (NS) iteration is much faster than that of SVD, but the total time of two methods is comparable. The authors of improved B-CNN  also proposed two other implementations, i.e.,  FP by NS iteration plus BP by SVD and FP by SVD plus BP by Lyapunov (Lyap.), which take 15.31 (2.09) and 12.21 (11.19), respectively. We observe that, in any case, the forward$+$backward time taken by  single meta-layer of improved B-CNN is significant  as GPU unfriendly SVD or EIG cannot be avoided, even though the forward computation is very  efficient when NS iteration is used. 
\begin{table}[t]
	\setlength\tabcolsep{4pt}
	\renewcommand{\baselinestretch}{1.05}
	\footnotesize
	\centering
	\begin{minipage}[t]{1.0\linewidth}
		\centering
		\begin{tabular}{|l|c|c|c|}
			\hline
			Method & Top-1 Error   &  Top-5 Error  & Time \\
			\hline
			\hline
			AlexNet~\cite{Krizhevsky2012ImageNet}       &  41.8     & 19.2  &  1.32 (0.77) \\
			\hline
			MPN-COV~\cite{Li_2017_ICCV}                           &  38.51 & 17.60 & 3.89 (2.59)\\
			B-CNN~\cite{lin2015bilinear}          &  39.89 & 18.32 & 1.92 (0.83)\\
			DeepO$_{2}$P~\cite{Ionescu_2015_ICCV} &  42.16 & 19.62  & 11.23 (7.04)\\
			Impro. B-CNN$^{*}$\cite{lin2017improved}     & 40.75         &  18.91  &  15.48 (13.04)  \\
			G$^2$DeNet~\cite{Wang_2017_CVPR}      &    38.71      & 17.66     &  9.86 (5.88) \\
			\hline
			\hline
			iSQRT-COV(Frob.)   &  38.78 &  17.67  & 2.56 (0.81)\\
			iSQRT-COV(trace) &  \textbf{38.45} &  \textbf{17.52} & 2.55 (0.81) \\
			\hline
		\end{tabular}
	\end{minipage}
	\renewcommand{\baselinestretch}{1.0}
	\caption{Error rate (\%) and time of FP+BP (ms) per image of different covariance pooling methods with AlexNet on ImageNet. Numbers in parentheses indicate FP time.  $^{*}$Following~\cite{lin2017improved}, improved B-CNN  successively  performs matrix square root, element-wise square root and $\ell_{2}$ normalizations.}
	\label{table:second-order-AlexNet}
\end{table}
Tab.~\ref{table:time-comparison}\subref{table:time-matrix-decomposition}
presents  running time  of EIG and SVD of an $256\times 256$ covariance matrix.   Matlab~(M) built-in CPU functions and GPU functions deliver over 10x and 2.1x  speedups over their CUDA counterparts, respectively.  Our method needs to store $\mathbf{Y}_{k}$ and $\mathbf{Z}_{k}$ in Eqn.~(\ref{equ:coupled-equation}) which will be used in backpropagation, taking up more memory than EIG or SVD based ones. Among all,  our iSQRT-COV ($N=5$) takes up the largest memory of 1.129MB, which is insignificant compared to 12GB memory on a Titan Xp. Note that for network inference only, our method  takes  0.125MB memory as it is unnecessary to store  $\mathbf{Y}_{k}$ and $\mathbf{Z}_{k}$.

Next, we compare in  Fig.~\ref{fig:Comparison_time} speed  of network training between MPN-COV and iSQRT-COV with both one-GPU and two-GPU configurations. For one-GPU configuration, the speed gap vs. batch size between the two methods keeps nearly constant. For two-GPU configuration, their speed gap becomes more significant when  batch size gets larger. As can be seen, the  speed of iSQRT-COV network continuously grows with increase of batch size while that of MPN-COV tends to saturate when batch size is larger than 512. Clearly our iSQRT-COV network can make better use of computing power of multiple GPUs than MPN-COV.

\vspace{4pt}\noindent\textbf{Pre-normalization by Trace vs. by Frobenius Norm}\quad 
Sec.~\ref{section:proposed-method} describes two pre-normalization methods. Here we compare them in Tab.~\ref{table:second-order-AlexNet} (bottom rows), where iSQRT-COV (trace) indicates pre-normalization by trace. We can see that pre-normalization by trace produces 0.3\% lower error rate than that by Frobenius norm, while taking similar time with the latter. Hence, in all the remaining experiments, we adopt trace based pre-normalization method.

\vspace{4pt}\noindent\textbf{Comparison with Other Covariance Pooling Methods}\quad 
We compare iSQRT-COV  with other covariance pooling methods, as shown in Tab.~\ref{table:second-order-AlexNet}. The results of MPN-COV, B-CNN and DeepO$_{2}$P are duplicated from~\cite{Li_2017_ICCV}. We train from scratch  G$^2$DeNet and improved B-CNN  on ImageNet. We use the most efficient implementation  of improved B-CNN, i.e., FP by SVD and BP by Lyap., and we mention all implementations of improved B-CNN produce similar results. Our iSQRT-COV using pre-normalization by trace is marginally better than  MPN-COV.  All  matrix square root normalization methods  except improved B-CNN outperform  B-CNN and DeepO$_{2}$P. Since  improved B-CNN is identical to MPN-COV if element-wise square root  normalization and $\ell_{2}-$normalization are neglected, 
\begin{table}[t]
	\setlength\tabcolsep{4pt}
	\renewcommand{\baselinestretch}{1.05}
	\footnotesize
	\centering
	\begin{minipage}[t]{1.0\linewidth}
		\centering
		\begin{tabular}{|c|l|c|c|}
			\hline
			Pre-normalization & Post-compensation & Top-1 Err.   &  Top-5 Err. \\
			\hline
			\hline
			\multirow{3}{*}{Trace} & $\quad\;$w/o       &   N/A    & N/A     \\
			& $\quad\;$w/~~BN~\cite{DBLP:journals/corr/IoffeS15}       &  23.12     & 6.60     \\
			& $\quad\;$w/~~Trace        &  $\textbf{22.14}$ & $\textbf{6.22}$ \\
			\hline
		\end{tabular}
	\end{minipage}
	\renewcommand{\baselinestretch}{1.0}
	\caption{Impact of post-compensation on iSQRT-COV  with  ResNet-50 architecture on ImageNet.}
	\label{table:Post-compensation}
\end{table}
\begin{figure}[t]
	\setlength\tabcolsep{4pt}
	\footnotesize
	\centering
	\begin{minipage}[b]{0.7\linewidth}
		\centering
		\includegraphics[width=1.0\textwidth]{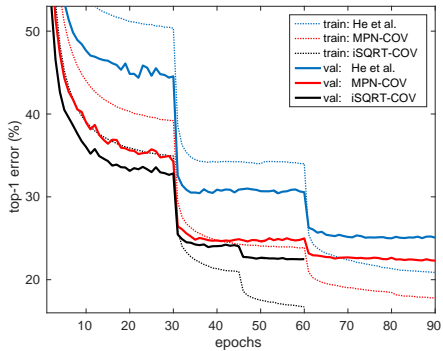}
	\end{minipage}
	\caption{Convergence curves of different networks trained  with ResNet-50 architecture on ImageNet.}
	\label{figure:traing-iSQRT-COV}
\end{figure}
its unsatisfactory performance  suggests that,  after matrix square root normalization, further element-wise square root normalization and $\ell_{2}-$normalization hurt large-scale ImageNet classification. This is consistent with the observation in~\cite[Tab. 1]{Li_2017_ICCV}, where after matrix power normalization,  additional normalization by Frobenius norm or matrix $\ell_{2}-$norm makes performance decline.

\subsection{Results  on ImageNet  with ResNet Architecture}\label{section:ImageNet-ResNet}

This section evaluates  iSQRT-COV with  ResNet architecture~\cite{He_2016_CVPR}. We  follow~\cite{Li_2017_ICCV} for color augmentation and weight initialization. We rescale each training image  with its shorter side randomly sampled on $[256, 512]$~\cite{Simonyan15}. The fixed-size $224\times224$ patch is randomly cropped from the rescaled image or its horizontal flip. We rescale each test image with a shorter side of 256 and evaluate a single $224\times 224$ center crop for inference. We use SGD with a mini-batch size of 256, a weight decay  of 0.0001 and a momentum of 0.9. We train  iSQRT-COV networks from scratch in 60 epochs, initializing    the learning rate  to $10^{-1.1}$ which is divided by 10 at epoch 30 and 45, respectively. 

\vspace{4pt}\noindent\textbf{Significance  of Post-compensation}\quad 
Rather than our post-compensation scheme, one may choose Batch Normalization (BN)~\cite{DBLP:journals/corr/IoffeS15} or simply do nothing (i.e., without post-compensation). \begin{table}[thb]
	\setlength\tabcolsep{4pt}
	\renewcommand{\baselinestretch}{1.05}
	\footnotesize
	\centering
	\begin{minipage}[t]{1.0\linewidth}
		\centering
		\begin{tabular}{|l|c|c|c|}
			\hline
			Method & Model & Top-1 Err.   &  Top-5 Err. \\
			\hline
			\hline
			\parbox{0.80in}{ \vspace{2pt}He et al.~\cite{He_2016_CVPR}} & \multirow{5}{*}{ResNet-50} &  24.7 &  7.8 \\
			\parbox{0.80in}{ \vspace{2pt}FBN~\cite{LiYanghao_2017_ICCV}}&  & 24.0 & 7.1 \\
			\parbox{0.80in}{ \vspace{2pt}SORT~\cite{Wang_2017_ICCV}}  &   & 23.82 & 6.72  \\
			\parbox{0.80in}{ \vspace{2pt}MPN-COV~\cite{Li_2017_ICCV}}  &   & 22.73  & 6.54 \\
			\parbox{0.90in}{ \vspace{2pt}iSQRT-COV\vspace{2pt}}  &   &  \textbf{22.14}   &  \textbf{6.22} \\
			\hline
			\hline
			\parbox{0.80in}{\vspace{2pt}He et al.~\cite{He_2016_CVPR}} & \multirow{2}{*}{ResNet-101} & 23.6  & 7.1 \\
			\parbox{0.90in}{ \vspace{2pt}iSQRT-COV \vspace{2pt}}  &   &  \textbf{21.21}   &  \textbf{5.68} \\
			\hline
			\hline
			\parbox{0.80in}{\vspace{2pt}He et al.~\cite{He_2016_CVPR}} & ResNet-152 & 23.0  & 6.7 \\
			\hline
		\end{tabular}
	\end{minipage}
	\renewcommand{\baselinestretch}{1.0}
	\caption{Error (\%) comparison of  second-order networks with first-order ones on ImageNet.}
	\label{table:ImageNet-ResNet}
\end{table}
Tab.~\ref{table:Post-compensation} summarizes impact of different schemes on iSQRT-COV network with ResNet-50 architecture. Without post-compensation, iSQRT-COV network fails to converge. Careful observations show that in this case  the gradients are very small (on the order of $10^{-5}$), and largely tuning of learning rate helps little. Option of BN helps the network converge, but producing about 1\% higher top-1  error rate than our post-compensation scheme. The comparison above suggests that our post-compensation scheme is essential for achieving state-of-the-art results.

\vspace{-8pt}\paragraph{Fast Convergence of iSQRT-COV Network}
We compare convergence of iSQRT-COV and MPN-COV with ResNet-50 architecture, as well as the original ResNet-50~\cite{He_2016_CVPR} in which  global average pooling is performed after the last convolution layer. Fig.~\ref{figure:traing-iSQRT-COV} presents the convergence curves. Compared to the original ResNet-50, the convergence of both iSQRT-COV and MPN-COV is significantly faster. We  observe that  iSQRT-COV can converge well within 60 epochs, achieving top-1 error rate of 22.14\%, $\sim$0.6\% lower than MPN-COV. We also trained iSQRT-COV with  90 epochs using  same setting with MPN-COV, obtaining top-5 error of 6.12\%, slightly lower than  that with 60 epochs (6.22\%). This indicates  iSQRT-COV can converge in less epochs, so further accelerating training, as opposed to MPN-COV.   The fast convergence property of iSQRT-COV is appealing. As far as we know, previous networks with ResNet-50 architecture require at least 90 epochs to converge to  competitive results.

\vspace{4pt}\noindent\textbf{Comparison with State-of-the-arts}\quad 
In Tab.~\ref{table:ImageNet-ResNet}, we compare our method with other second-order networks, as well as the original ResNets. With ResNet-50 architecture, all the second-order networks improve over the first-order one while our method performing best. MPN-COV and iSQRT-COV, both of which involve square root normalization, are superior to  FBN~\cite{LiYanghao_2017_ICCV} which uses no normalization and SORT~\cite{Wang_2017_ICCV} which introduces dot product transform in the linear sum of two-branch module followed by element-wise normalization. Moreover, our iSQRT-COV  outperforms MPN-COV by  0.6\% in top-1 error. Note that our 50-layer iSQRT-COV network achieves lower error rate  than much deeper  ResNet-101 and ResNet-152, while our 101-layer iSQRT-COV network outperforming the original ResNet-101 by 2.4\% and ResNet-152 by 1.8\%, respectively.


\vspace{4pt}\noindent\textbf{Why Approximate Square Root Performs Better}\quad  Fig.~\ref{fig:impact_iteration-N} shows that  more iterations which lead to more accurate square root is not helpful for  iSQRT-COV with AlexNet. 
\begin{table}[thb]
	\setlength\tabcolsep{4pt}
	\renewcommand{\baselinestretch}{1.05}
	\footnotesize
	\centering
	\begin{minipage}[t]{1.0\linewidth}
		\centering
		\begin{tabular}{|l|c|c|c|c|c|}
			\hline
			Method  & $d$ & Dim. & Top-1 Err.   &  Top-5 Err. & Time \\
			\hline
			\hline
			He et al.~\cite{He_2016_CVPR}  & N/A & 2K& 24.7 &  7.8 & 8.08 (1.93) \\
			\hline
			\multirow{3}{*}{iSQRT-COV}   & 64 & 2K    & 23.73   & 6.99 &  9.86 (2.39) \\
			&  128 & 8K        & 22.78  & 6.43  & 10.75 (2.67)\\
			&  \;256\; & 32K       & 22.14  & 6.22 & 11.33 (2.89)\\
			\hline
		\end{tabular}
	\end{minipage}
	\renewcommand{\baselinestretch}{1.0}
	\caption{Error rate (\%) and time of FP+BP (ms) per image vs. $d$ (or representation  dimension) of compact  iSQRT-COV with ResNet-50  on ImageNet. Numbers in parentheses indicate FP time.  }
	\label{table:compact-iSQRT-COV}
\end{table}
From Tab.~\ref{table:ImageNet-ResNet}, we observe that iSQRT-COV with ResNet computing approximate square root performs better than MPN-COV which can obtain exact square root by EIG. Recall that, for covariance pooling ConvNets, we face the problem of small sample of large dimensionality,  and matrix square root is consistent with general shrinkage principle of robust covariance estimation~\cite{Li_2017_ICCV}. Hence,  we conjuncture that approximate  matrix square root  may be a better robust covariance estimator than the exact square root. Despite this analysis, we think this  problem is  worth future research.

\vspace{4pt}\noindent\textbf{Compactness of iSQRT-COV}\quad 
Our iSQRT-COV outputs  32k-dimensional  representation which is high. Here we consider to compress this representation. Compactness by PCA~\cite{lin2015bilinear} is not viable since obtaining the principal components on ImageNet is too expensive. CBP~\cite{Gao_2016_CVPR} is not applicable to our iSQRT-COV as well, as it does not \textit{explicitly} estimate the covariance matrix. We propose a simple scheme, which decreases the dimension (dim.)  of covariance representation by lowering the number $d$ of channels of $1\times 1$ convolutional layer before our covariance pooling.  
Tab.~\ref{table:compact-iSQRT-COV} summarizes  results of compact iSQRT-COV. The recognition error increases slightly  ($\uparrow 0.64\%$) when $d$ decreases from 256 to 128 (correspondingly, dim. of image representation $32\text{K}\to 8\text{K}$). The error rate is 23.73$\%$ if the dimension is compressed to 2$\text{K}$, still outperforming the original ResNet-50 which performs global average pooling.

\subsection{Fine-grained Visual Categorization (FGVC)}

Finally, we apply  iSQRT-COV models pre-trained on ImageNet to FGVC. 
For fair comparison, we follow~\cite{lin2015bilinear} for experimental setting and evaluation protocol. On all  datasets, we crop $448\times 448$ patches as input images.  We replace  1000-way softmax layer of a pre-trained iSQRT-COV model by a k-way softmax layer, where $k$ is number of classes in the fine-grained dataset, and finetune the network using SGD with momentum of 0.9 for 50$\sim$100 epochs with a small learning rate ($lr$$=$$10^{-2.1}$) for all layers except the fully-connected layer, which is set to $5\times lr$. We use horizontal flipping as data augmentation. After finetuning, the outputs of iSQRT-COV layer are  $\ell_{2}-$normalized before inputted to train $k$ one-vs-all linear SVMs with hyperparameter $C=1$. We predict the label of a test image  by averaging SVM scores of the  image and its horizontal flip.   

\begin{table}[thb]
\setlength\tabcolsep{6pt}
\renewcommand{\baselinestretch}{1.05}
\footnotesize
\centering
\begin{minipage}[t]{1.0\linewidth}
\centering
\begin{tabular}{|c|c|c|c|c|c|}
\hline
                                     & Method  & Dim. & Birds   &  Aircrafts  & Cars \\
\hline
\hline
 \multirow{4}{*}{ \rotatebox{90}{{ResNet-50}}} & \multirow{2}{*}{\parbox{0.58in}{\centering \vspace{2pt}iSQRT-COV}}  
                                                      & 32K  & $\textbf{88.1}$  & $\textbf{90.0}$   & $ \textbf{92.8}$\\
  \cline{3-6}
                                                  &  &   8K    & 87.3    & 89.5      &  91.7\\
     \cline{2-2}\cline{3-6}
& CBP~\cite{Gao_2016_CVPR}                            & 14K    & 81.6    & 81.6      &  88.6\\
\cline{2-6}
& KP~\cite{Cui_2017_CVPR}                              & 14K   & 84.7    & 85.7      &  91.1\\
\hline 
\hline
\multirow{6}{*}{ \rotatebox{90}{{VGG-D}}}& iSQRT-COV &  32K   & 
{87.2}  & {90.0}      &  {92.5}\\
\cline{2-6}
&NetVLAD~\cite{Arandjelovic_2016_CVPR}                          &  32K    & 81.9  & 81.8      &  88.6\\
\cline{2-6}
&CBP~\cite{Gao_2016_CVPR}                         &  8K    & 84.3  & 84.1      &  91.2\\
\cline{2-6}
& KP~\cite{Cui_2017_CVPR}                          &  13K    & 86.2  & 86.9      &  92.4\\
\cline{2-6}
& LRBP~\cite{Kong_Charless_2017_CVPR}                          &  10K    & 84.2  & 87.3      &  90.9\\
\cline{2-6}
  &\parbox{0.55in}{\centering \vspace{2pt}Improved\\ B-CNN\cite{lin2017improved}}     & 262K  & 85.8  & 88.5      &  92.0\\
 \cline{2-6}
& G$^2$DeNet~\cite{Wang_2017_CVPR}                 & 263K & 87.1  & 89.0      &  {92.5}\\
\cline{2-6}
& HIHCA~\cite{Cai_2017_ICCV}                       &   9K   & 85.3  & 88.3      &  91.7 \\
\hline
\hline
\multicolumn{2}{|l|}{iSQRT-COV with ResNet-101}     &  32K  & \color{blue}{\textbf{88.7}}  & \color{blue}{\textbf{91.4}}      &  \color{blue}{\textbf{93.3}}  \\
\hline
\end{tabular}
\end{minipage}
\renewcommand{\baselinestretch}{1.0}
\caption{Comparison of accuracy (\%) on fine-grained benchmarks. Our method uses neither bounding boxes nor part annotations.}
\label{table:Fine-grained}
\end{table}

Tab.~\ref{table:Fine-grained} presents classification results of different methods, where column 3 lists the dimension of the corresponding representation. 
With ResNet-50 architecture, KP  performs much better than CBP, while  iSQRT-COV (8K) respectively outperforms KP (14K) by about 2.6\%, 3.8\% and 0.6\% on Birds, Aircrafts and Cars,  and iSQRT-COV (32K) further improves accuracy. Note that  KP combines first-order up to fourth-order statistics while iSQRT-COV only exploits second-order one.  With VGG-D, iSQRT-COV (32k)
matches or outperforms  state-of-the-art competitors, but  inferior to iSQRT-COV (32k) with ResNet-50.

On  all fine-grained datasets, KP and CBP with 16-layer VGG-D  perform  better than their counterparts with 50-layer ResNet, despite the fact that  ResNet-50 significantly  outperforms VGG-D  on ImageNet~\cite{He_2016_CVPR}.  The reason may be that the last convolution layer of pre-trained ResNet-50 outputs 2048-dimensional features, much higher than 512-dimensional one of VGG-D, which are not suitable for  existing second- or higher-order pooling methods. \textit{Different from all existing methods which use models pre-trained on ImageNet with first-order information, our pre-trained models are of second-order.} Using pre-trained iSQRT-COV models with ResNet-50, we achieve recognition results superior to all the compared methods, and furthermore, establish state-of-the-art results on three fine-grained benchmarks using iSQRT-COV model with ResNet-101.

\section{Conclusion}

We presented an iterative matrix square root normalization of covariance pooling (iSQRT-COV)  network which  can be trained end-to-end. Compared to existing works depending heavily on GPU unfriendly EIG or SVD, our method, based on coupled  Newton-Schulz iteration~\cite{Higham:2008:FM}, runs much faster as it involves only matrix multiplications, suitable for parallel implementation on GPU. We validated our method on both large-scale ImageNet dataset and challenging fine-grained benchmarks. Given efficiency and promising performance of our iSQRT-COV, we hope    global covariance pooling  will be a promising alternative to  global average pooling in other deep network architectures, e.g., ResNeXt~\cite{Xie_Ross_2017_CVPR}, Inception~\cite{DBLP:journals/corr/IoffeS15} and DenseNet~\cite{Huang_2017_CVPR}.

{\small
\bibliographystyle{ieee}
\bibliography{egbib}
}
\end{document}